\documentclass[conference]{IEEEtran}

\usepackage{lineno,hyperref}
\usepackage{graphicx}
\usepackage{float}
\usepackage{xcolor}

\begin{document}
%\begin{frontmatter}
%\title{IoT Enabled Healthcare Analytics}
\title{Multi-Class and Automated Tweet Categorization}


\author{
\IEEEauthorblockN{Khubaib Ahmed Qureshi}
\IEEEauthorblockA{DHA Suffa University\\
DG-78, Off Khayaban-e-Tufail, \\Ph-VII (Ext.),
DHA, Karachi-75500, Pakistan\\
Email: k.ahmed@dsu.edu.pk}
}
%\IEEEauthorblockA{Karachi Institute of Economics and Technology
%PAF Base, Korangi Creek, Karachi, Pakistan}
%\and
%\IEEEauthorblockN{Muhammad Taha Jilani}
%\IEEEauthorblockA{Karachi Institute of Economics and Technology
%PAF Base, Korangi Creek, Karachi, Pakistan Email: m.taha@pafkiet.edu.pk}
%\and
%\IEEEauthorblockN{Khurum Nazir Junejo}
%\IEEEauthorblockA{Ibex CX, Pakistan Email: %junejo@gmail.com}
%}

\maketitle

\begin{abstract}
Twitter is among the most prevalent social media platform being used by millions of people all over the world. It is used to express ideas and opinions about political, social, business, sports, health, religion, and various other categories. The study reported here aims to detect the tweet category from its text. It becomes quite challenging when text consist of 140 characters only, with full of noise. The tweet is categorized under 12 specified categories using Text Mining or Natural Language Processing (NLP), and Machine Learning (ML) techniques. It is observed that huge number of trending topics are provided by twitter but it is really challenging to find out that what these trending topics are all about. Therefore, it is extremely crucial to automatically categorize the tweets into general categories for plenty of information extraction tasks. A large dataset is constructed by combining two different nature of datasets having varying level of category identification complexities. It is annotated by experts under proper guidelines for increased quality and high agreement values. It makes the proposed model quite robust. Various types of ML algorithms were used to train and evaluate the proposed model. These models are explored over three dataset separately. It is explored that the nature of dataset is highly non-linear therefore complex or non-linear model perform better. The best ensemble model named, Gradient Boosting achieved an AUC score of 85\%. That is much better than the other related studies conducted.  
\end{abstract}

%\end{frontmatter}

%\linenumbers

%\end{IEEEkeywords}

% \fi
\IEEEpeerreviewmaketitle

\section{Introduction}
Today social media is becoming a need, every day thousands of new users get registered on various social media platforms like Facebook, Twitter, etc. These platforms are used to express emotions, thoughts, creative ideas, raising voice on social issues, and many more things. Twitter is a social media platform in which users can post textual messages and visual messages which are called as tweets.  Tweets can also contain hashtags like \#Entertainment, \#Religion, etc. to express the topic of the message. Users can also be tagged in a tweet by mentioning the user e.g. @Ibrahim. Twitter also provides a list of trending hashtags based on the user's country. These trending topics need to be further processed for extracting the meaning because they don't comprehend much for user. Due to the large number of messages posted on Twitter, there is a need for an automated approach to extract key information or simple categories from these short messages. The automated tweet category detection techniques has various important applications related to information extraction. It serves as necessary building block for intelligent information extraction and processing. That may include topic-based sentiment analysis, topic experts identification, genre-based microblogs search engines, topic-based recommendation systems, etc. \\
There are various tweet categorization detection techniques reported in the literature but they are conducted on the limited dataset with normal accuracy and are only useful for single or few category detection like politics \cite{1COTELO201654}, election \cite{2tare2014multi}. The study reported in \cite{3vijayaraghavan2016automatic} works on three categories like politics, sports, and technology. The semantic model developed to identify six tweet topics on limited dataset of 1330 tweets \cite{2ibtihel2018semantic}. There is a another related work that have experimented only Naive Bayes, SVM and Decision Tree models \cite{1lee2011twitter} and build their tweet categorization model over trending topics. \\
The model presented in our study is unique in three respects: first, it can predict the topic of tweet with several categories like Entertainment, Social Issues, Well-being, News, Sports, Religion, etc. They cover all discourse. Second, it is based on large dataset of varying nature of topic identification complexity. Specially the Twitter Top 50 User dataset contains maximum informality contents like net-speaks, fillers, short forms, assent, non-fluencies, abusive language, smilies, etc. It is annotated by 12 experts and 4 reviewers. Annotation was done under proper guidelines to enhance inter-agreement rate and provide high quality. The dataset was also explored and found that its non-linear in nature that requires complex models for category classification. Third, it has experimented range of ML models. The proposed approach has important applications in marketing, recommender systems, credibility applications, etc.\\
Given the limited text of the tweet, it is very challenging to process and identify the topics/categories of the tweet. In our model, the key information is extracted from the tweet using ML and NLP approaches \cite{46113240}. We have used range of ML classifiers like Naive Bayes, Logistic Regression (LR), Support Vector Classifier (SVC), Random Forest, Muli-Layer Perceptron (MLP) and Gradient Boosting for the classification of tweets. These ML algorithms cover linear, non-linear, tree-based, non-tree-based, probabilistic, non-probabilistic, and bagging ensembles and boosting ensembles. It helps us to explore the proposed model behaviour over the datasets. Once the topic is detected then we can identify the expert domain, user reading interests, and various other attributes of any user. \\
The rest of the papers is organized as section \ref{sec:ds} describes the process of data gathering, quantity of data, and assigning categories to data. Section 3 describes the methodology and text processing framework in addition to data pre-processing techniques that include dropping of invalid data, data splitting, encoding, stemming, and lemmatization. Section 4 presents features extraction and exploration, followed by section 5 that describes the ML models used in the paper. The conclusion and future research directions are discussed in Section 6 and 7 respectively.

\begin{figure}[b]
\centering
\includegraphics[width=1.0\linewidth]{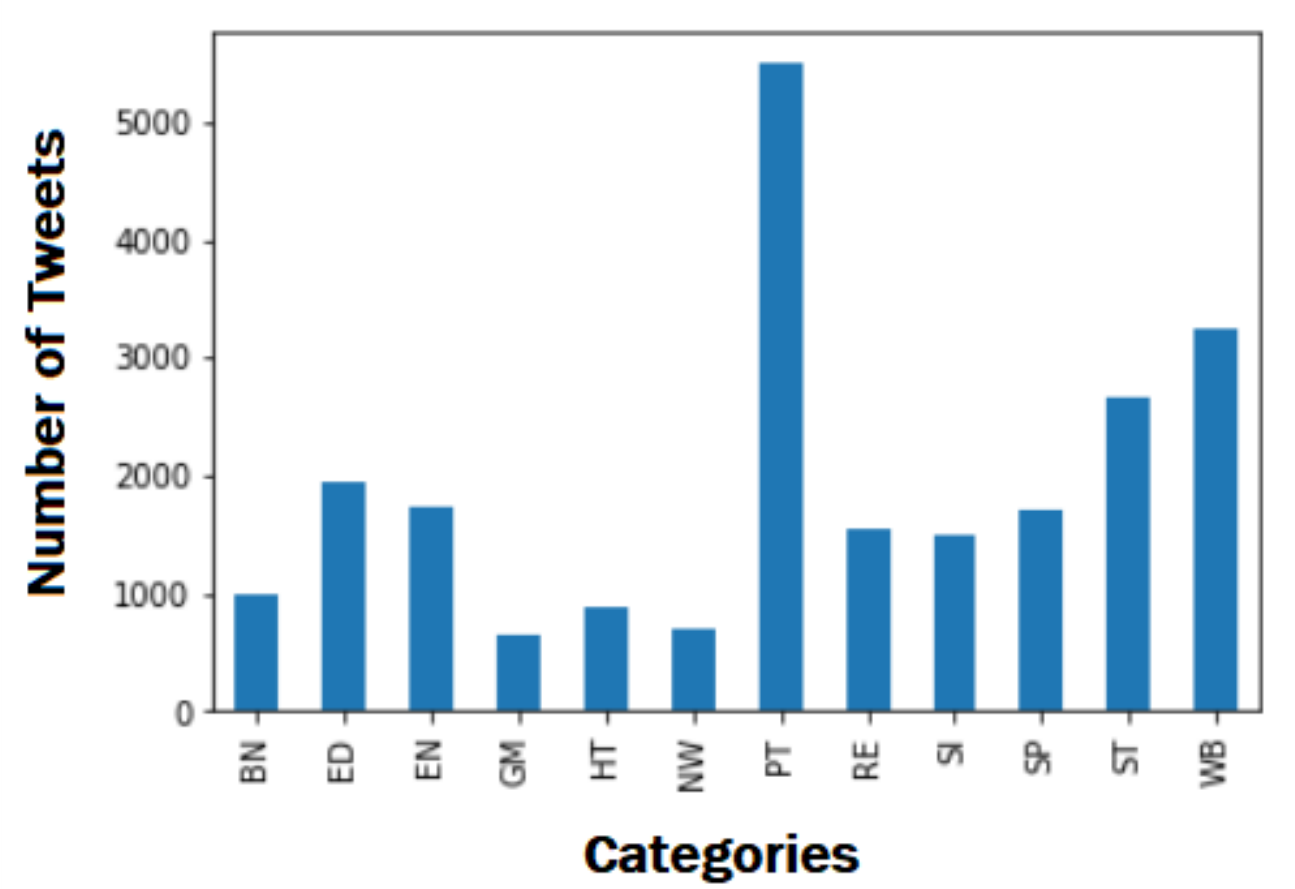}
\caption{Number of Tweets against each category in Top Users Tweet Data}
\label{fig:pic1}
\end{figure}

\begin{figure}[b]
\centering
\includegraphics[width=1.0\linewidth]{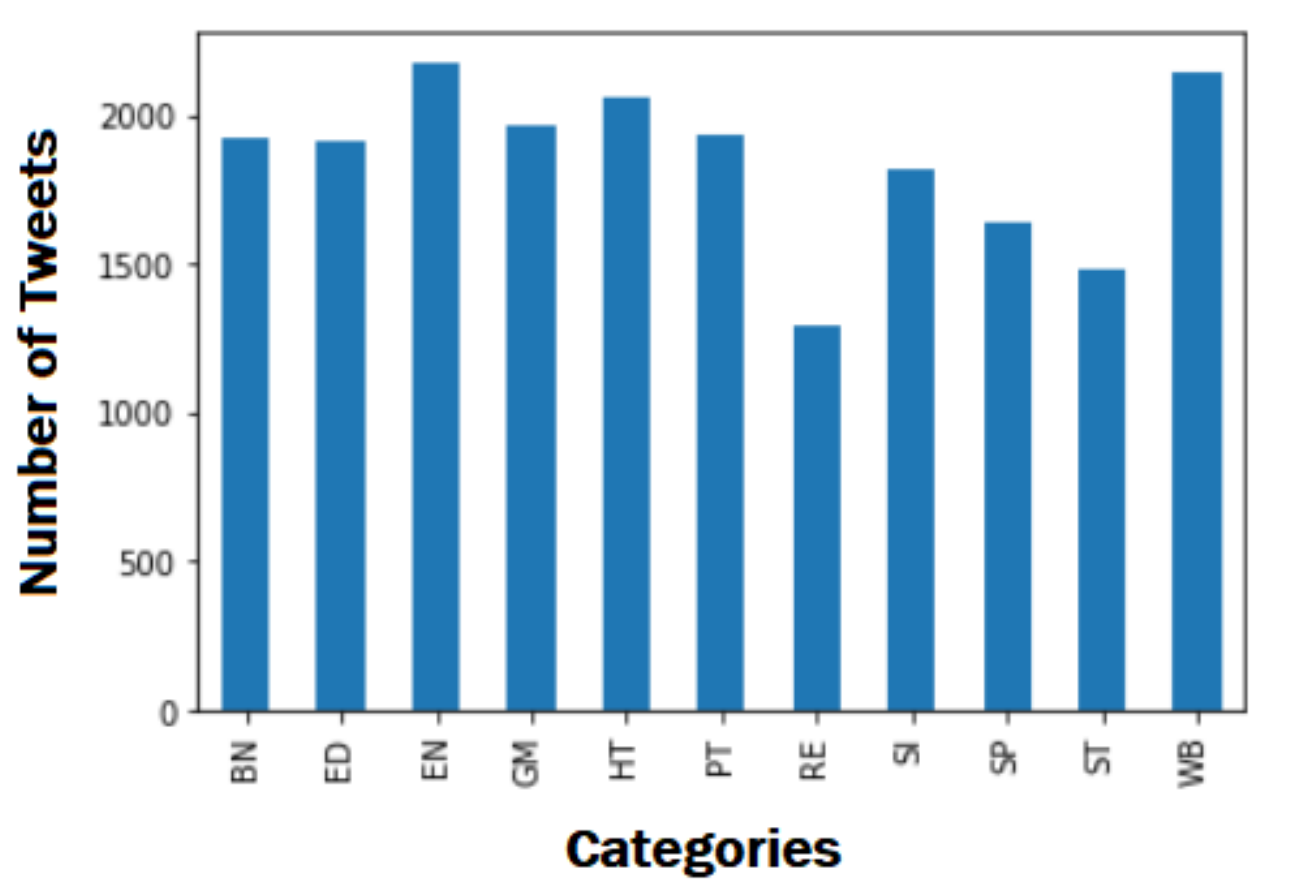}
\caption{Number of Tweets against each category in Synonym Based Tweet Data}
\label{fig:pic2}
\end{figure}

\begin{figure}[b]
\centering
\includegraphics[width=1.0\linewidth]{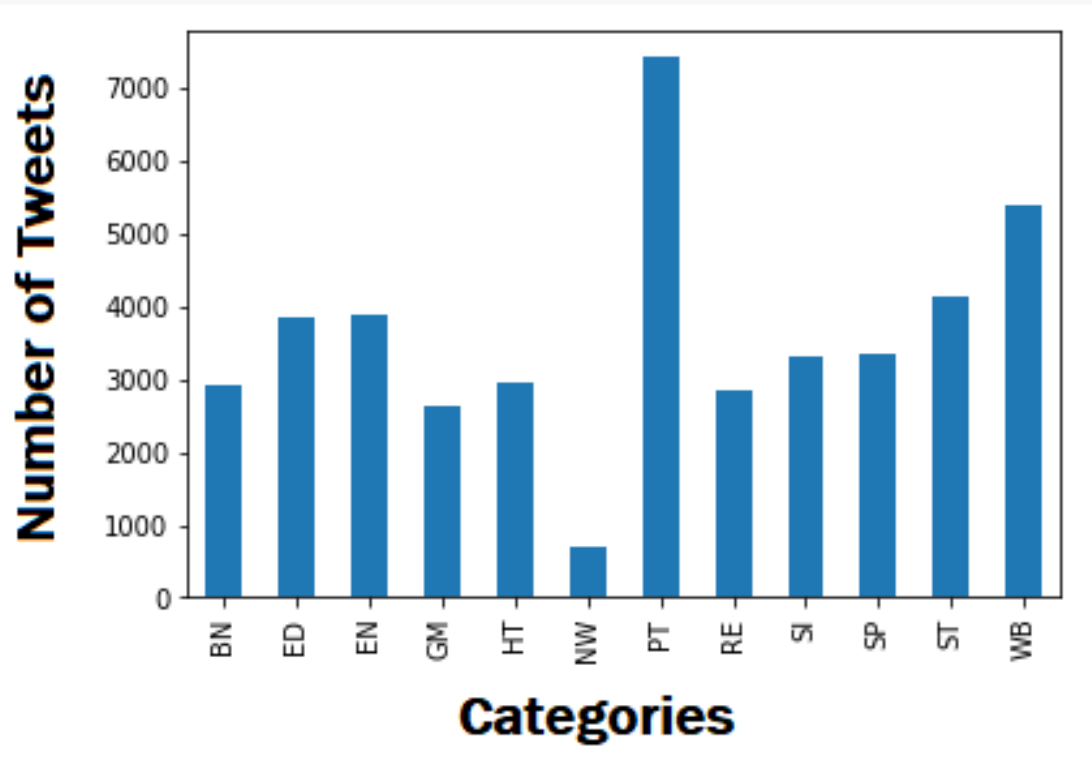}
\caption{Number of Tweets against each category in Combined Data}
\label{fig:pic3}
\end{figure}

\section{Data Collection}
\label{sec:ds}
Two datasets have been used in the proposed study. One is 'Twitter Top 50 Users Tweet Data' that contains 23090 tweets while the other one is 'Synonym Based Tweet Data' that contains 20338 tweets. Both of these datasets are further described in the following subsections.

\subsection{Top Users Tweet Data}
The data was collected using Twitter API "tweepy" \cite{5roesslein2019tweepy}. We particularly fetch data of persons who were famous in the following categories mentioned in the Table \ref{tab:cat}. They are known as Top Twitter Users with respect to top 50 topics. The timeline tweets from these top 50 users were fetched and then group them with respect to our specified categories. The dataset generated from such user's timelines are extremely complex because their language is completely informal and full of slang, containing short forms, abusive language, Net-speaks, Fillers, Assents, etc. The total number of tweets collected for the dataset was 23090. Among these tweets, 17317 tweets were used for training the model while 5773 tweets were used for testing the model.

\begin{table}[tbh]

\centering
\caption{Categories for Top Users Tweet Data}

\label{tab:cat}

\begin{tabular}{|l|}
\hline
\textbf{Categories} \\ \hline \hline
SCIENCE \& TECHNOLOGY (ST) \\
POLITICS (PT) \\
HEALTH (HT) \\
BUSINESS (BN) \\
EDUCATION (ED) \\
SPORTS (SP) \\
ENTERTAINMENT (EN) \\ 
SOCIAL ISSUES (SI) \\
RELIGION (RE) \\ 
GENERAL ADMIN \& MANAGEMENT (GM) \\
WELL BEING (WB) \\ 
NEWS  (NW) \\
REJECT  (RJ) \\ \hline
\end{tabular}

\end{table}

\subsection{Synonym Based Tweet Data}
\begin{figure*}[tbh]
\centering
\includegraphics[width=0.5\linewidth,height=5cm]{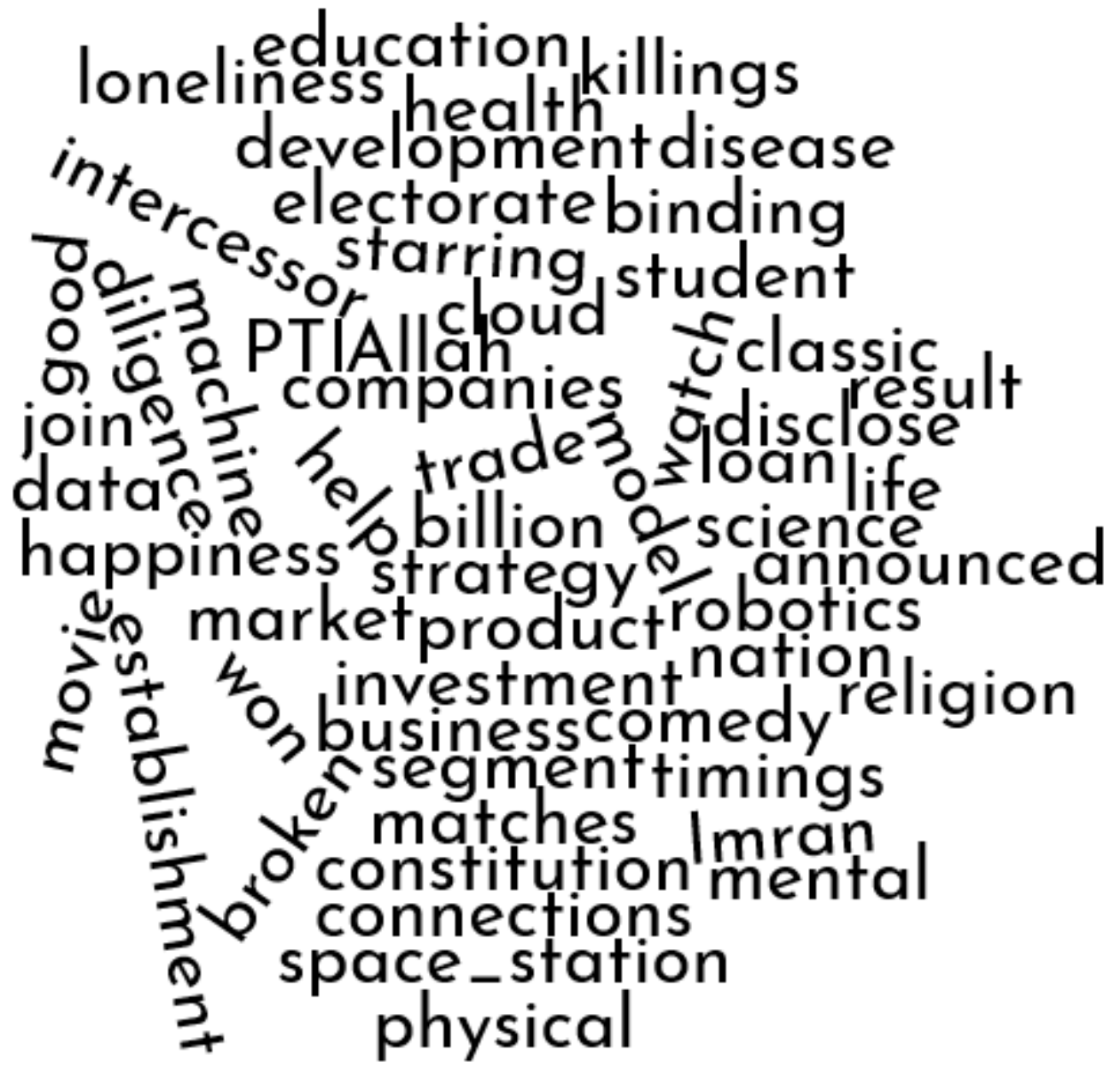}
\caption{Frequent Words}
\label{fig:pic7}
\end{figure*}

Python Library ‘GetOldTweets3’ \cite{12GetOldTweets3} was used to fetch tweets to increase the size of the dataset. In the beginning, 50 tweets for a specific category like health were fetched. These tweets were then passed into the model. Through the model, we acquired the largest coefficient contained by the tweets. Taking this coefficient (see figure \ref{fig:pic7} as sample for top frequent words for the categories) as a synonym for the category we fetched more accurate tweets related to the corresponding category. This process was repeated for all eleven categories. In this way, we got accurate and relevant data for each chosen category. The entire set of categories were fetched to get 20338 tweets. Among these 15253 tweets were used for training the model while 5085 tweets were used to test the model.\\
Finally, we combined these two datasets to create a large dataset of 43428 tweets. The nature of both datasets were completely different. The synonym based dataset has simple and easy to understand English while top-user dataset has slang-based, much informal and quite complex or difficult to understand language. This was done due to construct the model most realistic and versatile in topic identification.
%%%%%%%%%%%%%%%
subsection{Data Annotation}  
\begin{table*}[tbh]
\centering
\caption{Topic Tagging Guidelines}
\label{tab:topictagging}
%\resizebox{4cm}{14cm}{
\begin{tabular}{p{1.5in}p{5in}}
\hline
\textbf{Topic} & \textbf{Description}\\ \hline \hline
SCIENCE \& TECHNOLOGY (ST) & It includes CS, IT, medical, Astronomy, Physics, Chemistry, Mathematics, Bio, all Engineering related contents (training \& education, invention, fictions, etc.)\\

POLITICS (PT) 
& Self explanatory\\

HEALTH (HT) & It includes treatments, fitness, ailments, diagnosis, symptoms, physiotherapy, medicines, health related precautions, Dr’s advice, related to living things, etc.\\
BUSINESS (PROFESSIONS, ECONOMICS, FINANCE, and MARKETING) (BN) & It includes all professions, sale purchase, sale offering, admissions, services offered, cost/rates, and all marketing stuff\\
EDUCATION (ED)& It includes things like schools/university/class, teaching, learning methodologies, exams, and all other subjects not included in Science \& Tech e.g.: English, Urdu, Arts, law, humanities, etc.\\
SPORTS (SP)& Self explanatory\\
ENTERTAINMENT (EN)& It includes stuff related to movies, media, music, picnic, parties, traveling, hoteling, love \& affection enjoyment, jokes \& comedy, drama, story telling, etc.\\
SOCIAL ISSUES (SI)& It includes very broad range of issues and their discussions related to poverty, governance, justice, law and order, transportation, marriage, economy, inflation, security, job crisis, human rights, etc.\\
RELIGION (RE)& Self explanatory\\
GENERAL ADMIN \& MANAGEMENT (GM)& It includes all announcements, notices, reminders, SOP’s, rules/regulation/policies and their dissemination e.g.: some event/class/seminar/workshop/conference canceled, closed, postponed, open, schedules, as well as any lost and stolen notice even human, etc.\\
WELL BEING (WB)& It includes all quotes related to ethics, wisdom, life, etc. any general stuff for the betterment of humanity e.g.: some traffic jam observed, some trap and risk observed, and related recommendations to mass audience for betterment, similarly charity and donations, all society social work, etc.\\
NEWS (NW)& It only includes stuff from news channels (not treated as category but a group/source, further divided in all above categories) therefore tag as: NW - XX\\
REJECT (RJ)& all others\\ \hline
\end{tabular}
%}

\end{table*}
\begin{figure*}[h!]
\centering
\includegraphics[width=1.0\linewidth, height=10cm] {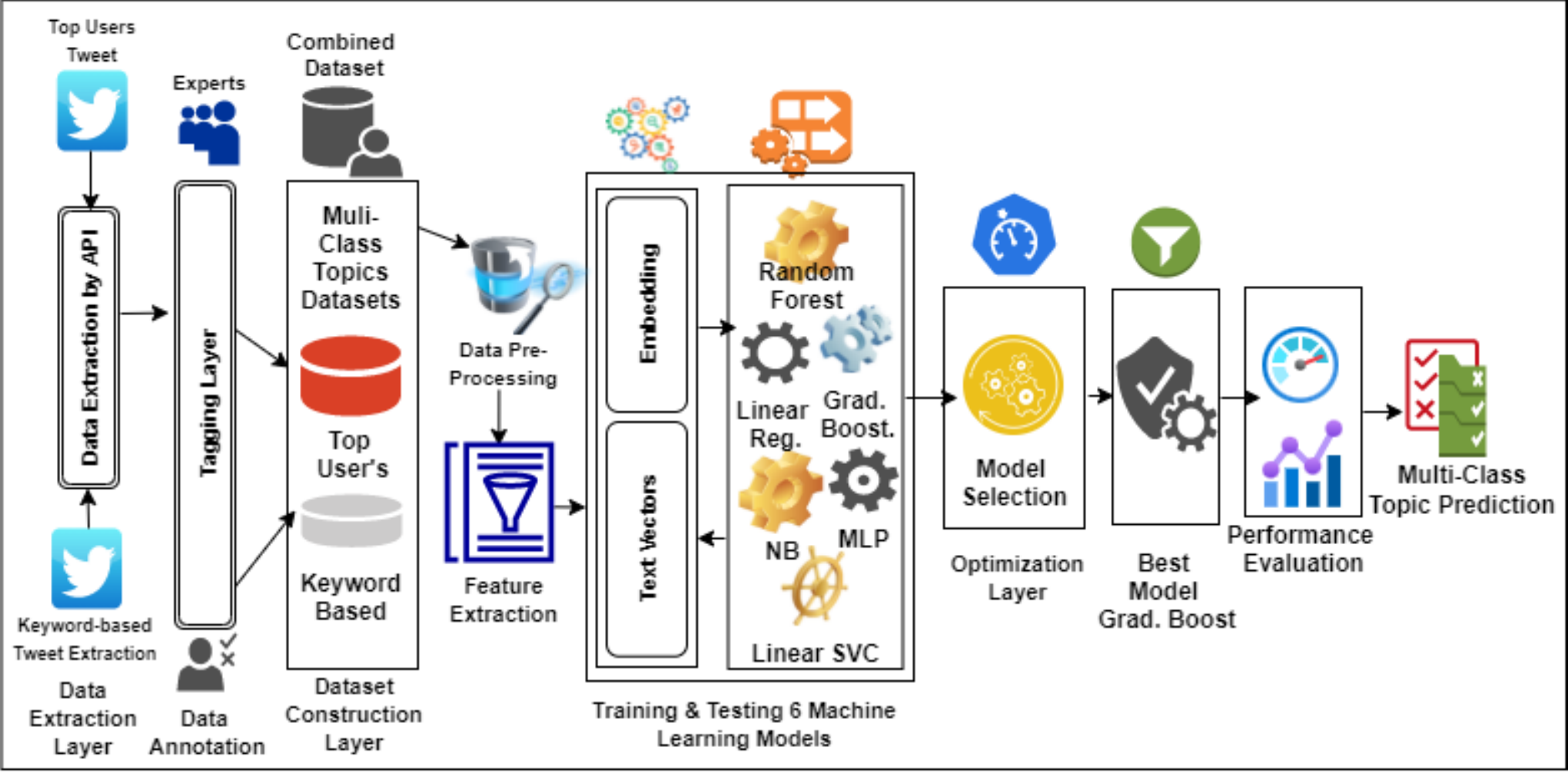}
\caption{High Level System Framework: complete processing from data extraction to prediction is shown, including complete machine learning pipeline. }
\label{mlf}
\end{figure*} 
%%%%%%%%%%%%%%%%%%
\begin{figure}[h!]
\centering
\includegraphics[width=1.0\linewidth, height=16.5cm] {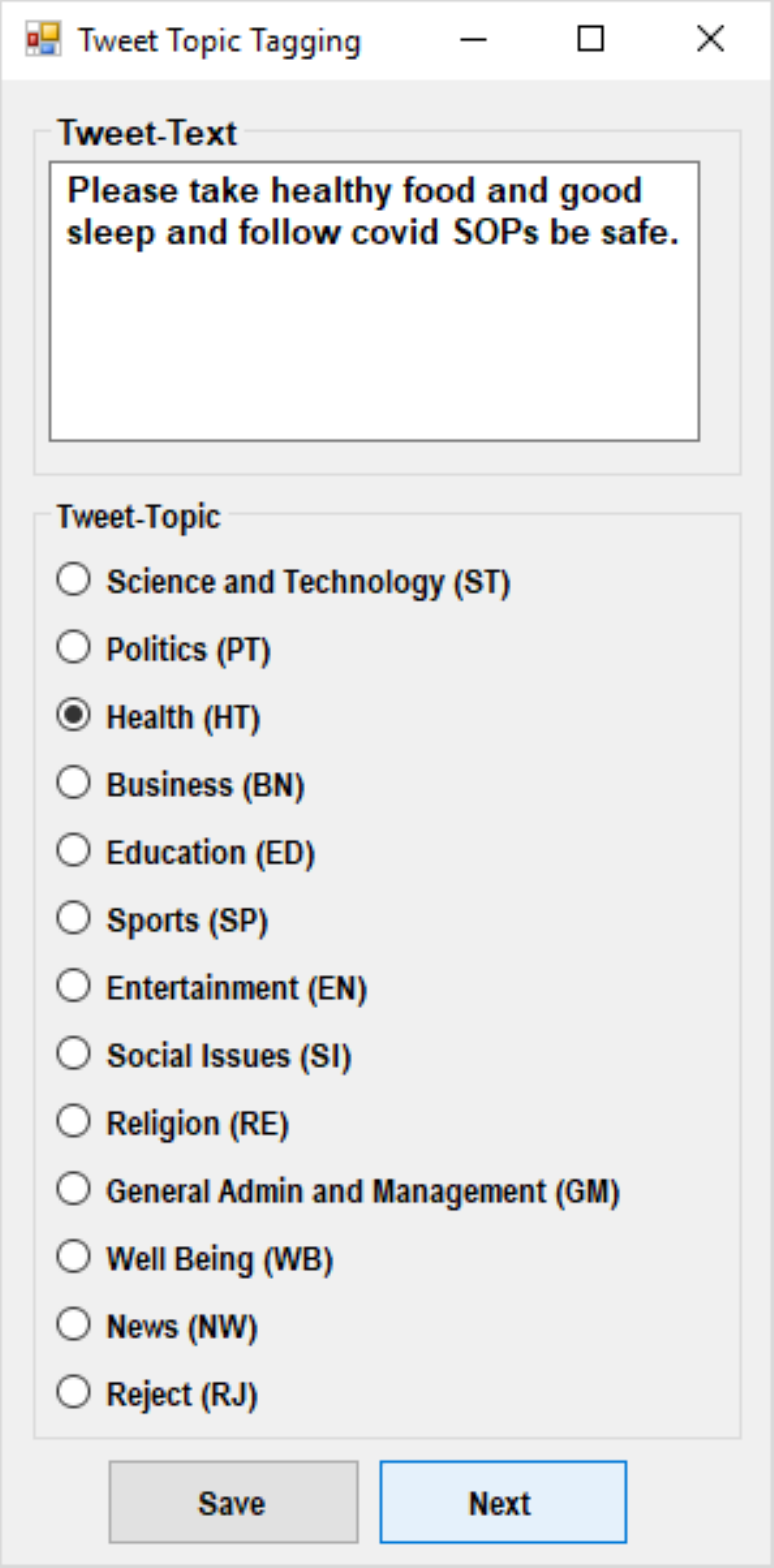}
\caption{Tweet Topic Tagging GUI}
\label{ml}
\end{figure} 
%%%%%%%%%%%%%%%
Data tagging is done through 12 experts. The taging was also reviewed by 4 reviewers. Making use of experts and reviewers eliminated the need of testing the annotators performance using gold-standard data. The annotation was done manually using custom developed GUI based application (see figure \ref{ml}). These experts were given the data of 43428 tweets for tagging. It took 6 months for developing the ground truth (GT). There were total 12 topics for tweet annotations that are given in table \ref{tab:cat}. The annotators were supported with the annotation application (see application's GUI in figure \ref{ml}). It shows all these topics to the annotators for selection. The annotators were furnished with complete guidelines and definitions that are given in table \ref{tab:topictagging} for enhancing the quality of data tagging and increasing the inter rater agreement. Total 12 experts were divided in 3 teams having 4 experts in each, therefore each post/tweet was tagged by 4 experts. The majority voting was used to develop the ground truth/label. There were three unique cases identified to develop the ground truth.  These ground truth cases were:  GT4YES, GT3YES, and GT2YES.\\
\textbf{GT4YES:} If a tweet topic is selected by all four annotators then that tweet is assumed YES for that topic.\\ 
\textbf{GT3YES:} If a tweet topic is selected by three annotators out of four, even then that tweet is assumed YES for that topic.\\
\textbf{GT2YES:}If a tweet topic is selected by any two annotators out of four, then that tweet is assumed YES for that topic. It is the case when there might be some different topics selected by other two annotators.\\
%%%%%%%%%%%%%%%%
\begin{table*}[h!]
\caption{Agreement (O) and Kappa (K) values between experts and ground truths.}
\label{kapa}
\begin{center}
\resizebox{16.2cm}{2.7cm}{%
\begin{tabular}{|l|c|c|c|c|c|c|c|}
\hline
  \textbf{Overlap(O) \& Kappa(K)} & \textbf{GT2YES(O)} & \textbf{GT2YES(K)} & \textbf{GT4YES(O)} & \textbf{GT4YES(K)} & \textbf{GT3YES(O)} & \textbf{GT3YES(K)} \\  
\hline
 SCIENCE \& TECHNOLOGY (ST) & 0.88 & 0.70 & 0.80 & 0.60 & 0.82 & 0.71 \\ 
\hline
 POLITICS (PT) & 0.98 & 0.69 & 0.88 & 0.65 & 0.95 & 0.67   \\
\hline
 HEALTH (HT) & 0.96 & 0.70 & 0.89 &0.64 & 0.94 & 0.66 \\
\hline
 BUSINESS (BN) & 0.98 & 0.77 & 0.84 & 0.66 & 0.90 & 0.70  \\
\hline
 EDUCATION (ED)  & 0.86 & 0.68 & 0.80 & 0.60 & 0.82 & 0.64\\  
\hline
 SPORTS (SP) & 0.96 & 0.76 & 0.88 & 0.70 & 0.92 & 0.74 \\  
\hline
 ENTERTAINMENT (EN) & 0.85 & 0.60 & 0.76 & 0.56 & 0.80 & 0.58  \\ 
\hline
SOCIAL ISSUES (SI) & 0.88 & 0.70 & 0.80 & 0.66 & 0.82 & 0.68   \\
\hline
 RELIGION (RE) & 0.98 & 0.80 & 0.90 & 0.75 & 0.94 & 0.78  \\ 
\hline
 GENERAL ADMIN \& MANAGEMENT (GM) & 0.90 & 0.68 & 0.85 & 0.64 & 0.88 & 0.66  \\ 
 \hline
 WELL BEING (WB) & 0.95 & 0.74 & 0.82 & 0.69 & 0.86 & 0.70  \\ 
  \hline
 NEWS (NW) & 0.86 & 0.60 & 0.80 & 0.57 & 0.82 & 0.58  \\ 
 
\hline
\end{tabular}%
}
\end{center}
\end{table*} 
\\
Considering the table \ref{kapa} that provides the overlap and the Cohen's Kappa values. The overlap is represented with "O" and Cohen's Kappa is represented with "K". The overlap measures the overlap between GTs and expert answers while Kappa provides the measure based on assumption that the overlap is occurred by some coincidence. Referring the table \ref{kapa}, one can observe that considering the overlap values the GT2YES provides the highest agreement. It could also be observed that for Kappa the best values are given by same GT2YES. It is quite evident that best values are obtained by GT2YES for both the overlap and Kappa therefore GT2YES will be used for finalizing the tweets labels. One can observe that some topics have achieved comparatively low values due to inherent complexity found in their tag identification, such topics are Science and Technology, Education, Entertainment, Social Issues and News. Remaining all topics were found strong in their label identification. 
%%%%%%%%%%%%%%%%
\section{Methodology}
The tweet topic detection is achieved by text processing framework that is implemented through following components (see high level framework components in figure \ref{mlf}). The data is extracted from twitter in extraction layer using keywords and from top-user timelines, then it was annotated in annotation layer. These annotated tweets from two sources, were combined in dataset construction layer. The dataset's tweets were pre-processed and their features were extracted using bag-of-word model (BOW) model. Different type of models were also exercised like: frequency vectorizer, TF-IDF Vectorizer, Weighted TF-IDF Vectorizer, OKAPI BM25, etc. In machine learning training and testing layer, the text vectors with their embeddings were constructed to get machibe learning (ML) models ready for experimentation. There were six ML models of different types that were evaluated. These models were better tuned in optimization layer and best model Gradient Boost were finally evaluated for multi-class classification.\\
The complete system framework is shown in figure \ref{ml}. The complete processing from data extraction to prediction is shown, including complete machine learning pipeline. 
Following are the steps of model development:

\subsection{Data Pre-Processing}
The data prepreparing step is very significant and crucial for managing content information without any loss of information. There are various techniques which make sure that the data obtained contains significant information. It includes removing invalid and dismissed qualities, stemming, and lemmatization.

\subsection{Dropping Invalid Data}
As our information is as an exceed expectations spreadsheet, we will join every one of the sheets and burden the information through pandas. We will quickly drop the pointless segments, for example, "id", "source" and "created at" as these won't be useful contributions for our models. In any case, we notice that there are two "tags" segments. The purpose behind this is in the dataset there is an invalid "tags" segment, which we will drop after combining the qualities with the right "tags" segment. Presently, our dataset just comprises of "tweets" and "tags" which are the features and labels. 
Next, we will utilize dropna() \cite{9pandas} to drop every one of the information missing or inadequate information from our data frame. After showing the information, we can see that there is a few miss-marked information with the labels or tags "ET", "RN",”WBB” which doesn't exist in our classes, so we will drop that information as it just contains insignificant information, alongside "RJ" and "Rj" which are dismissed lines for example the tweets which don't fall in a particular classification. 
At last, we have a perfect pandas data frame which is prepared or the subsequent stage of preprocessing.

\subsection{Data splitting and encoding}
Data is part into preparing, and testing set utilizing sklearn's model selection work with 0 irregular state, which parts it into a 75/25 proportion. Label Encoder is utilized from sklearn.preprocessing which encodes every one of our labels into numericqualities, so it is simpler to recognize for the machine. At last, CountVectorizer is applied to change our tweets into a check grid.

\subsection{Stemming and Lemmatization}
We are utilizing stemmer from gensim library \cite{8gensim} to preprocess the information. Right off the bat, stop words are removed and after that information is stemmed and tokenized through lemmatizer. The information is at that point annexed into a cluster lastly, we are changing over the string into a cluster. Presently our preprocessing is done. We have the accompanying two sections: 

Content - This section comprises of vectors of words which are tokenized and stemmed. 

\begin{figure}[tbh]
\centering
\includegraphics[width=1.0\linewidth]{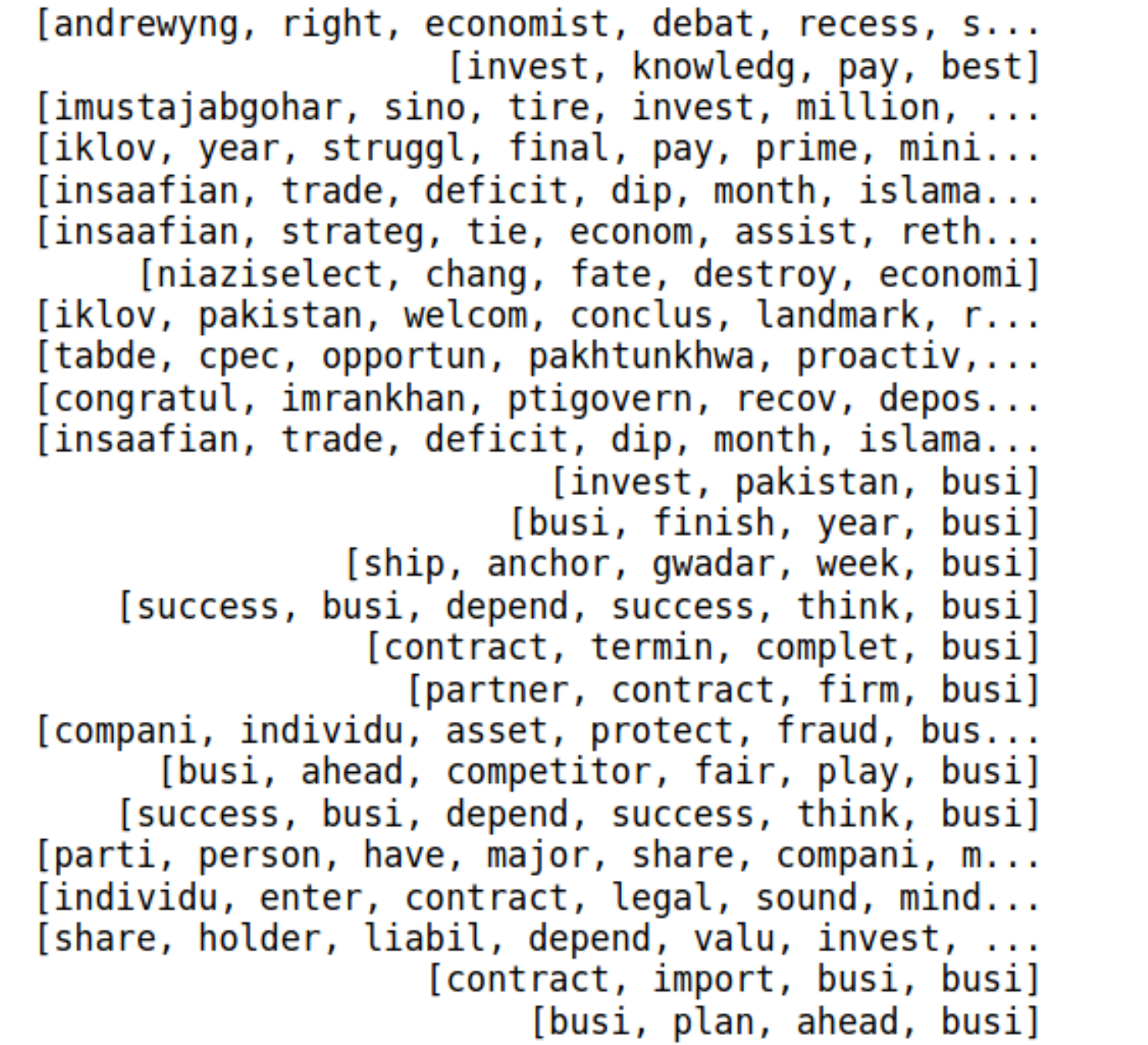}
\caption{Encoded Names}
\label{fig:pic8}
\end{figure}

Labels - This section comprises of encoded names as shown in figure \ref{fig:pic8}.

\begin{figure*}[tbh]
\centering
\includegraphics[width=1.0\linewidth]{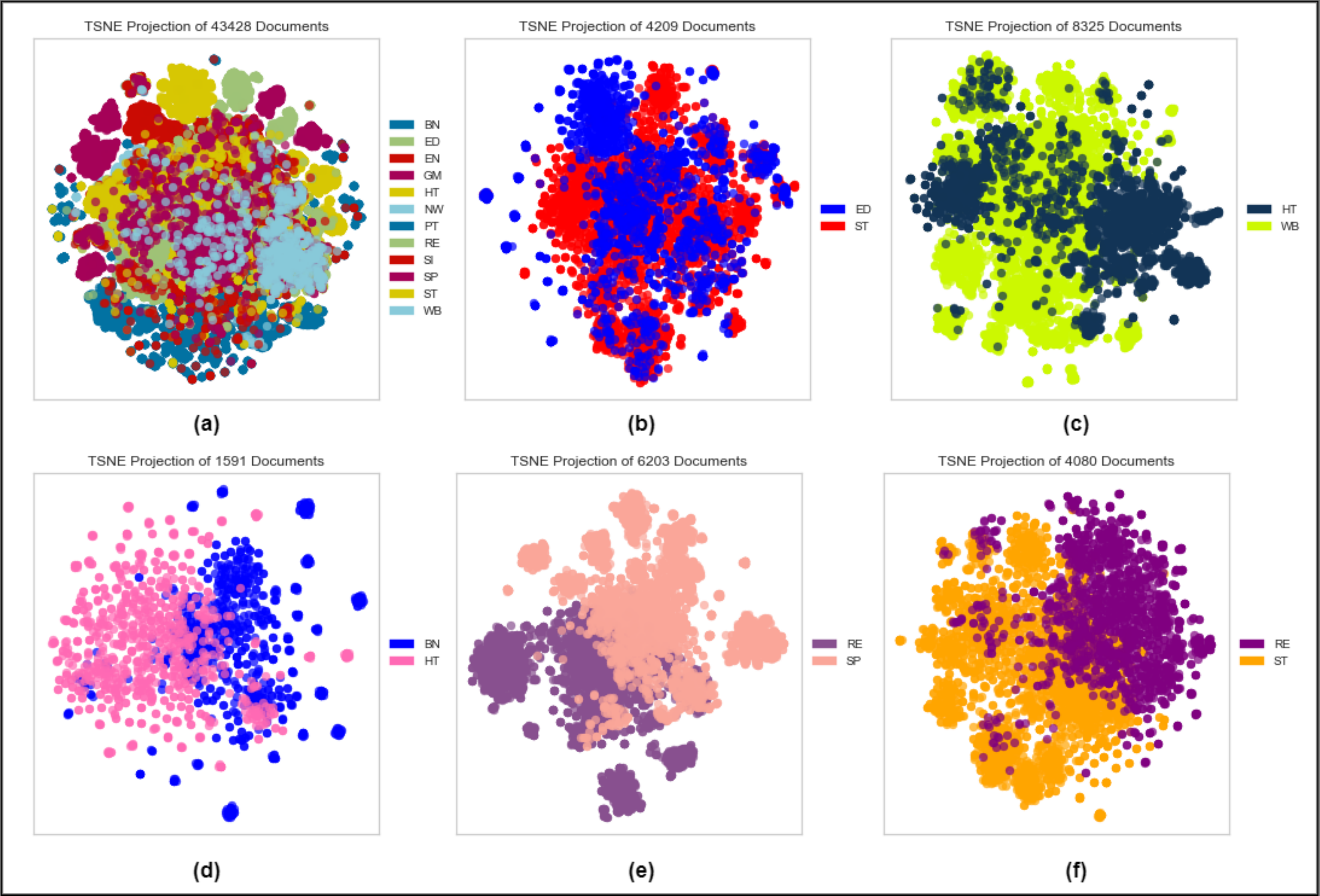}
\caption{Dataset Plotting using tSNE through Model Features: (a) All-Topic Plot. (b) Science \& Technology vs Education (c) Health vs Well-Being. (d) Business vs Health. (e) Religion vs Sports. (f) Religion vs Science and Technology. Cases (b) and (c) are examples of complex boundary cases, while cases (d)(e) and (f) are presenting simple or clearly separating boundary cases.}
\label{fig:TSNE}
\end{figure*}
\section{Feature Extraction and Exploration}
The pre-procession is done and the tweets are now completely ready for features extraction. Tweet features are extracted through Term-Tweet incidence matrix. It is constructed using the Count-Vectorizer and their embeddings are also developed. These embeddings were extremely useful that can identify the similar terms or synonyms which enhance the model performance. \\
These features were used to develop high-dimensional plots known as t-distributed stochastic neighbor embedding (tSNE). It is a high-dimensional plotting method for visualizing the multi-dimensional data by giving each data point a location in a 2D or 3D map \cite{tSNE}. The extracted features are used as dimensions and each post is a data point and the 2D plots are developed. These plots are presented in figure \ref{fig:TSNE}.\\
There are two different cases that are observed. First case presents those topics that are completely different from one another and therefore their identification is quiet easy and produced very good results. These are presented in figure \ref{fig:TSNE} (e) of Religion vs Sports, and figure \ref{fig:TSNE} (f) presenting Religion vs Science and Technology. The same case is followed in Business vs Health that is shown in figure \ref{fig:TSNE} (d). All these cases are clearly separating to one another. \\
The second case exist where two topics are having inherent overlaps and complex boundaries exist that are difficult to identify therefore more mis-classification reported. These cases could be examined in figures \ref{fig:TSNE} (b) and (c) between Science \& Technology vs Education and Health vs Well-Being, respectively. \\
The figure \ref{fig:TSNE} (a) shows all-topics tSNE plot. It represents that the topic detection problem is complex and the dataset shows non-linearity. It needs complex or non-linear models for their identification. The figure \ref{fig:TSNE} (a) also presents that many of the topics are clearly separating while only few are having mixed boundaries. One can observe that all these diagrams also present that our features are good enough to separate all these topics from one-another.   

\section{Models}
We trained our models with three data sets on six different machine learning algorithms. Following different models have been used with provided Hyper-parameters and best results are also shown in table \ref{tab:results}.

\subsection{Machine Learning Models for Top User’s Dataset}

\subsubsection{Naïve Bayes}
In the Naive base multinomialNB model, we took alpha = 0.1. The Naive Bayes is used because it is a good model when we have a small size of data set. It converges faster than the discriminative model and it is simple and easy to use.

\subsubsection{Linear SVC}
For Linear SVC we took C=0.1. The objective of a Linear SVC is to fit the model on the data provided by us. It returns the best-fitted hyperplane that classifies our data.

\subsubsection{Logistic Regression}
The parameter values for Logistic Regression were set as C=1.0, solver='lbfgs', multi\_class='multinomial'. Logistic Regression is used to describe the data and the relationship between one dependent binary variable and one or more nominal, ordinal, interval-independent variables. 

\subsubsection{Random Forest}
For Random Forest Classifier, we set the parameters as n\_estimators=500, max\_depth=200, random\_state=0. Random Forest is multiple decision trees. It merges all the results of the trees and predicts more accurately. It works like voting.

\subsubsection{Gradient Boosting}
For Gradient Boosting default parameters were used. Gradient Boosting is a machine learning technique for regression and classification problems. It produces a prediction model in the form of an ensemble of weak prediction models.

\subsubsection{Neural Network}
Neural Network’s model was trained with parameters as activation='relu', max\_iter=800, solver='lbfgs', learning\_rate\_init=0.005, hidden\_layer\_sizes=(46,44), random\_state=1. Neural Network has the ability to learn and model non-linear and complex relationship from data.

\subsubsection{Results}

\begin{figure*}[tbh]
\centering
\includegraphics[width=0.7\linewidth,height=7cm]{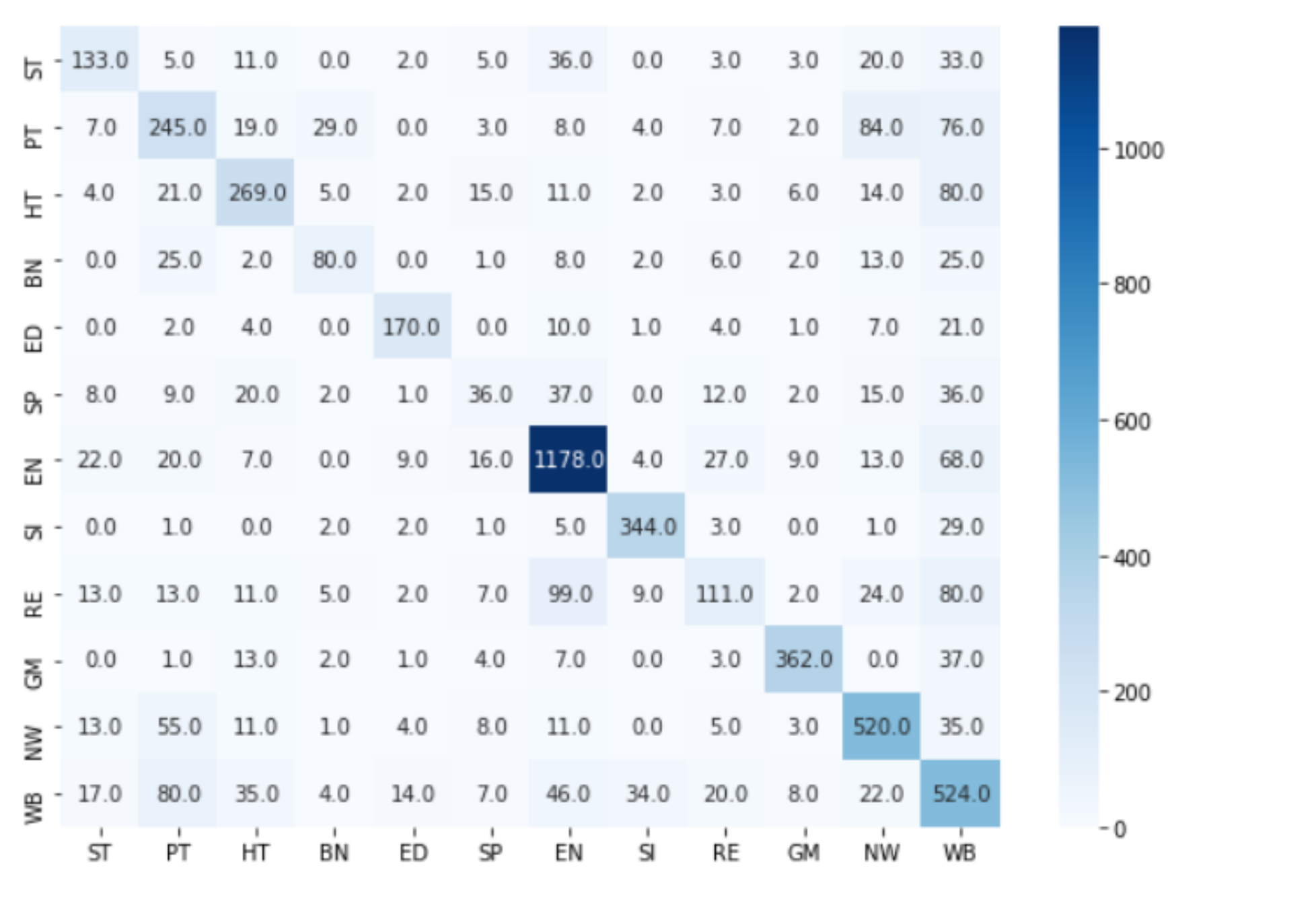}
\caption{Confusion Matrix for Top User’s Dataset}
\label{fig:pic4}
\end{figure*}

Considering the results of all the above algorithms conclude that all six models are highly over-fitted as shown in figure \ref{fig:pic4}. Considering the accuracy scores, the best result we found on this data set was using Gradient Boosting which gave 69.46\%. The other results that we found on this data set was using Neural Network having accuracy of 64.99\% and Naive Bayes having accuracy of 66.40\%. The Linear SVC produced accuracy of 68.82\% where as Logistic Regression scored 68.06\% accuracy and Random Forest performed 66.83\% accuracy. One can observe that AUC scores are much better for all these models. Since the dataset is imbalanced then AUC could be the better measure \cite{aucling2003auc}.

\begin{table*}[tbh]

\centering
\caption{Results of Machine Learning Models for Top User’s Dataset}

\label{tab:results}

\begin{tabular}{l|l|l|l|l|l}
\hline
\textbf{Model} & \textbf{Accuracy} & \textbf{Precision}& \textbf{Recall} & \textbf{F1} & \textbf{AUC}  \\ \hline \hline
Naive Bayes & 66.40\% & 0.61 & 0.63 & 0.62 & 0.79 \\
Linear SVC & 68.82\% & 0.66 & 0.62 & 0.63 & 0.79  \\
Logistic Regression & 68.06\% & 0.66 & 0.61 & 0.63 & 0.78 \\
Random Forest & 66.83\% & 0.68 & 0.57 & 0.59 & 0.76  \\
Neural Network & 64.99\% & 0.59 & 0.58 & 0.58 & 0.77  \\
Gradient Boosting & 69.46\% & 0.69 & 0.59 & 0.62 & 0.79  \\ \hline
\end{tabular}

\end{table*}

\begin{figure*}[tbh]
\centering
\includegraphics[width=0.7\linewidth,height=7cm]{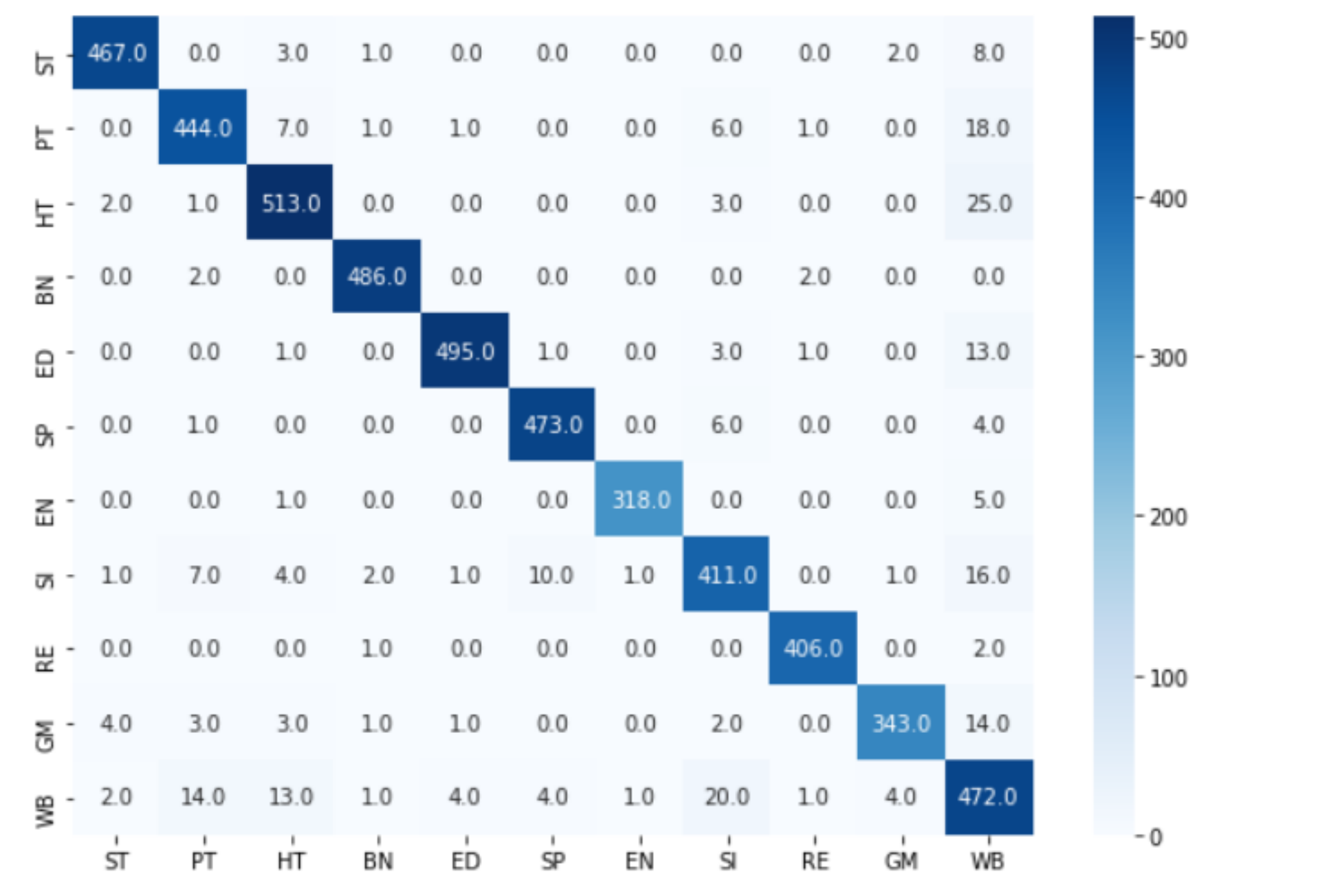}
\caption{Confusion Matrix for Synonym Based Dataset}
\label{fig:pic5}
\end{figure*}

\subsection{Machine Learning Models for Synonym Based Dataset}
The second experiment of training and testing models was carried out with a second dataset on all six algorithms used previously. The values of all the parameters were the same as done in the first dataset. All the algorithms performed well on the current data set. Considering the results of the algorithms conclude that these ML models are just-fit as shown in figure \ref{fig:pic5}. The best results were found using Gradient Boosting, and Random Forest, and Logistic Regression having accuracy of 96.01\%, 95.71\%, and 94.98\% respectively. The worst result was found using Naïve Bayes having accuracy of 89.30\%. A summary of the results of aforementioned models is described in table \ref{tab:results1}

\begin{table*}[tbh]

\centering
\caption{Results of Machine Learning Models for Synonym Based Dataset}

\label{tab:results1}

\begin{tabular}{l|l|l|l|l|l}
\hline
\textbf{Model} & \textbf{Accuracy} & \textbf{Precision}& \textbf{Recall} & \textbf{F1} & \textbf{AUC} \\ \hline \hline
Naive Bayes & 89.30\% & 0.90 & 0.90 & 0.90 & 0.94 \\
Linear SVC & 94.95\% & 0.95 & 0.95 & 0.95 & 0.97  \\
Logistic Regression & 94.98\% & 0.96 & 0.95 & 0.95 & 0.97 \\
Random Forest & 95.71\% & 0.96 & 0.96 & 0.96 & 0.97 \\
Neural Network & 93.90\% & 0.94 & 0.94 & 0.94 & 0.96  \\
Gradient Boosting & 96.01\% & 0.97 & 0.96 & 0.96 & 0.98  \\ \hline
\end{tabular}

\end{table*}

\subsection{Machine Learning Models for the Combined Dataset}
The third dataset was made by combining the first and second datasets. We merged both data sets Top tweet data set and Synonym Based data set and again trained the same machine learning models described above.

\subsubsection{Naïve Bayes}
We started Naive Bayes by setting the value of alpha as 2.6. For hyperparameter tuning of Naive Bayes, we use Randomized Search Cv. We did experimentation by changing the values of alpha from 2.0 to 3.0, in each iteration, we incremented it by 0.1.  It was found that the Naive Bayes algorithm performed well by setting the value of alphas as 2.6.

\subsubsection{Linear SVC} 
Linear SVC was tuned by looping hyperparameter on different values to find the best results. The model performed well by setting the hyperparameter value of C as 0.001.

\subsubsection{Logistic Regression}
For hyperparameter tuning of Logistic Regression, we used randomized Search Cv, parameter C, penalty, and solver. By iterating the model on different parameters values we got the value of C as 0.01. At this value, the logistic regression model performed well. After tuning we got the value of the solver as Lbfgs and the value of penalty as l2. 

\begin{figure*}[tbh]
\centering
\includegraphics[width=0.7\linewidth,height=7cm]{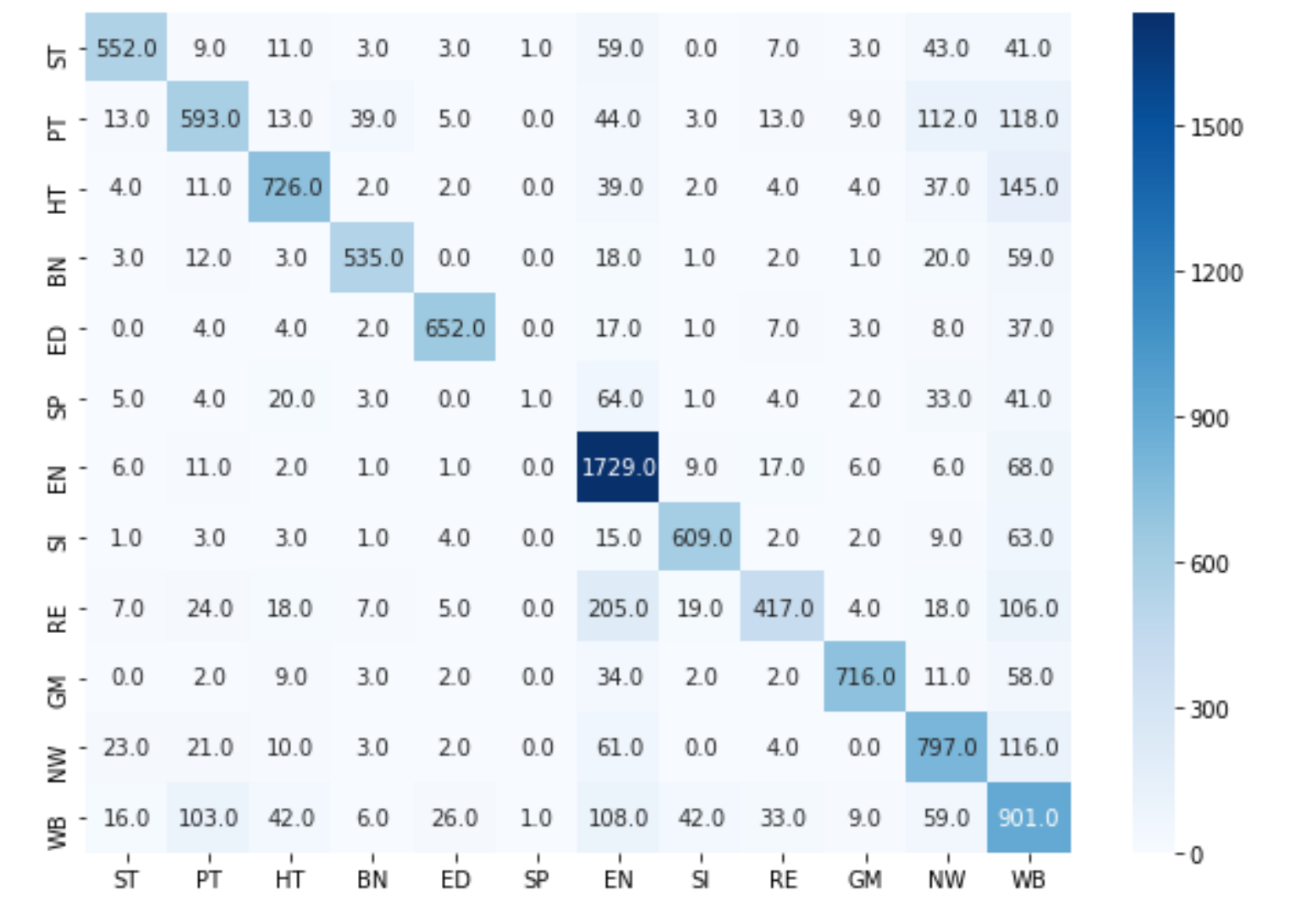}
\caption{Confusion Matrix of Gradient Boosting Model using Combined Dataset}
\label{fig:pic6}
\end{figure*}

\begin{table*}[tbh]
\centering
\caption{Results on Data set having training\_data size = 32571 and test\_data size = 10857}
\label{tab:results2}
\begin{tabular}{l|l|l|l|l|l}
\hline
\textbf{Model} & \textbf{Accuracy} & \textbf{Precision}& \textbf{Recall} & \textbf{F1} & \textbf{AUC} \\ \hline \hline
Naive Bayes & 76.10\% & 0.76 & 0.70 & 0.71 & 0.83 \\
Linear SVC & 76\% & 0.76 & 0.70 & 0.71 & 0.83 \\
Logistic Regression & 75\% & 0.76 & 0.69 & 0.71 & 0.83 \\
Random Forest & 75\% & 0.76 & 0.69 & 0.71 & 0.83  \\
Neural Network & 78\% & 0.74 & 0.73 & 0.73 & 0.80  \\
Gradient Boosting & 79\% & 0.80 & 0.74 & 0.75 & 0.85 \\ \hline
\end{tabular}

\end{table*}

\subsubsection{Random Forest}
For the tuning of Random Forest, we used randomized Search CV and checked different values of max-depth and estimator. We get the value of n-estimator as 400 and max-depth as 50. Our model performed well on the aforementioned values with the random state of 0.

\subsubsection{Gradient Boosting}
For Gradient Boosting default parameters were used.

\subsubsection{Neural Network}
For Neural Network MLPCLassifier we used Randomized Search Cv in which we tuned the model on the basis of parameters like max-iter, solver, learning-rate- init, batch-size, and alpha. We set them as max-iter = 100, solver=adam, learning-rate-init = 0.001, batch-size =128, and alpha =0.001.

\section{Conclusion}
In various information extraction tasks we need to identify the category of tweet to understand the what is being discussed or shared. It serves as building block for many intelligent applications. The tweet category identification becomes extremely challenging in micro-blog's short text messages hiving much noise. Therefore we have applied different data pre-processing techniques on our dataset before plugging in ML pipeline. Three dataset are constructed for analysis and explorations. These dataset were annotated by 12 experts and 4 reviewer with proper annotation guidelines that resulted in high agreement values between the annotators and showed good quality. The dataset would be available for community for further research and explorations. The final or combined dataset was explored through tSNE high dimensional plots. It is observed that the dataset is highly non-linear therefore non-linear models could solve the problem efficiently. We trained and evaluated six different machine learning models over our datasets like Naive Bayes \cite{6zhang2004optimality}, Linear SVC(support vector classifier), Logistic Regression, Random Forest classifier \cite{7Goel2017RandomFA}, Multi-Layer Perceptron (MLP), and Gradient Boosting models. Among them we get higher and better results on Gradient Boosting model that produced AUC of 0.85. The AUC score is much consistent specially for imbalanced data. The results are described in table \ref{tab:results2}. From the confusion matrix described in figure \ref{fig:pic6} we concluded that our model has predicted well results as humans also get confused whether text to be categorized as science and technology or education.

\section{Future Direction}
In the future, we decide to collect more data and add some new categories to our dataset. The inclusion of more data will allow us to apply Deep learning Models and to use Natural language processing to extract the topics from audio and picture files. We would also examine complex NLP features for the problem optimization.

\section{Acknowledgement}
Regarding the role of data extraction and annotation experts: Maad Saifuddin, Muhammad Mustajab Gohar, Shayan Mustafa, A. Kumail Pirzada, Ehtasham Ali, Syed Asad Hashmi, Ahsan Mukhtar, Ali Aslam, Muhammad Affan, Shahzad, Muhammad Shiraz, Masood Iqbal, Sheheryar Ali Azfar, and Areesh Ansari at Department  of  Computer Science, DHA Suffa University, Karachi, Pakistan, have extracted, labeled and compiled the topic datasets.

\bibliography{Security.bib}

\begin{thebibliography}{10}
\providecommand{\url}[1]{#1}
\csname url@samestyle\endcsname
\providecommand{\newblock}{\relax}
\providecommand{\bibinfo}[2]{#2}
\providecommand{\BIBentrySTDinterwordspacing}{\spaceskip=0pt\relax}
\providecommand{\BIBentryALTinterwordstretchfactor}{4}
\providecommand{\BIBentryALTinterwordspacing}{\spaceskip=\fontdimen2\font plus
\BIBentryALTinterwordstretchfactor\fontdimen3\font minus
  \fontdimen4\font\relax}
\providecommand{\BIBforeignlanguage}[2]{{%
\expandafter\ifx\csname l@#1\endcsname\relax
\typeout{** WARNING: IEEEtran.bst: No hyphenation pattern has been}%
\typeout{** loaded for the language `#1'. Using the pattern for}%
\typeout{** the default language instead.}%
\else
\language=\csname l@#1\endcsname
\fi
#2}}
\providecommand{\BIBdecl}{\relax}
\BIBdecl

\bibitem{1COTELO201654}
\BIBentryALTinterwordspacing
J.~Cotelo, F.~Cruz, F.~Enríquez, and J.~Troyano, ``Tweet categorization by
  combining content and structural knowledge,'' \emph{Information Fusion},
  vol.~31, pp. 54--64, 2016. [Online]. Available:
  \url{https://www.sciencedirect.com/science/article/pii/S1566253516000099}
\BIBentrySTDinterwordspacing

\bibitem{2tare2014multi}
M.~Tare, I.~Gohokar, J.~Sable, D.~Paratwar, and R.~Wajgi, ``Multi-class tweet
  categorization using map reduce paradigm,'' \emph{International Journal of
  Computer Trends and Technology (IJCTT)}, vol.~9, no.~2, pp. 78--81, 2014.

\bibitem{3vijayaraghavan2016automatic}
P.~Vijayaraghavan, S.~Vosoughi, and D.~Roy, ``Automatic detection and
  categorization of election-related tweets,'' in \emph{Tenth International
  AAAI Conference on Web and Social Media}, 2016.

\bibitem{2ibtihel2018semantic}
B.~L. Ibtihel, H.~Lobna, and B.~J. Maher, ``A semantic approach for tweet
  categorization,'' \emph{Procedia Computer Science}, vol. 126, pp. 335--344,
  2018.

\bibitem{1lee2011twitter}
K.~Lee, D.~Palsetia, R.~Narayanan, M.~M.~A. Patwary, A.~Agrawal, and
  A.~Choudhary, ``Twitter trending topic classification,'' in \emph{2011 IEEE
  11th International Conference on Data Mining Workshops}.\hskip 1em plus 0.5em
  minus 0.4em\relax IEEE, 2011, pp. 251--258.

\bibitem{46113240}
S.~Piao and J.~Whittle, ``A feasibility study on extracting twitter users'
  interests using nlp tools for serendipitous connections,'' in \emph{2011 IEEE
  Third International Conference on Privacy, Security, Risk and Trust and 2011
  IEEE Third International Conference on Social Computing}, 2011, pp. 910--915.

\bibitem{5roesslein2019tweepy}
J.~Roesslein, ``Tweepy documentation. 2009,'' \emph{Tweepy Documentation v3},
  vol.~5, 2019.

\bibitem{12GetOldTweets3}
D.~Mottl, ``Getoldtweets3 documentation release 0.0.11,'' November 2019,
  \url{https://pypi.org/project/GetOldTweets3/}.

\bibitem{9pandas}
W.~M. Team, ``Pandas: powerful python data analysis toolkit release 0.25.3,''
  November 2019,
  \url{https://pandas.pydata.org/pandas-docs/stable/reference/api/pandas.DataFrame.dropna.html}.

\bibitem{8gensim}
R.~R. rek, ``gensim documentation release 0.8.6,'' November 2017,
  \url{https://test-kek.readthedocs.io/_/downloads/en/stable/pdf/}.

\bibitem{tSNE}
L.~Van~der Maaten and G.~Hinton, ``Visualizing data using t-sne.''
  \emph{Journal of machine learning research}, vol.~9, no.~11, 2008.

\bibitem{aucling2003auc}
C.~X. Ling, J.~Huang, H.~Zhang \emph{et~al.}, ``Auc: a statistically consistent
  and more discriminating measure than accuracy,'' in \emph{Ijcai}, vol.~3,
  2003, pp. 519--524.

\bibitem{6zhang2004optimality}
H.~Zhang, ``The optimality of naive bayes,'' \emph{AA}, vol.~1, no.~2, p.~3,
  2004.

\bibitem{7Goel2017RandomFA}
E.~Goel and E.~Abhilasha, ``Random forest: A review,'' 2017.

\end{thebibliography}
\end{document}